\pdfoutput=1
\documentclass[10pt]{article}

\usepackage[letterpaper,margin=1in]{geometry}
\usepackage[T1]{fontenc}
\usepackage[utf8]{inputenc}
\usepackage{mathptmx}                 
\usepackage[scaled=0.9]{helvet}
\usepackage{courier}
\usepackage{microtype}
\usepackage{amsmath,amssymb}
\usepackage{booktabs}
\usepackage{array}
\usepackage[table]{xcolor}
\usepackage{tikz}
\usetikzlibrary{arrows.meta,positioning,fit,calc}
\usepackage[font=small,labelfont=bf,skip=6pt]{caption}
\usepackage{enumitem}
\usepackage[numbers,sort&compress]{natbib}
\usepackage[hidelinks]{hyperref}
\usepackage[capitalise,nameinlink]{cleveref}
\usepackage{url}
\usepackage{titlesec}
\titlespacing*{\paragraph}{0pt}{1.2ex plus .5ex minus .2ex}{0.8em}

\hypersetup{
  pdftitle    = {Deep Reinforcement Learning for Equity Trading: Benchmarking Actor-Critic Methods with Forward Retraining},
  pdfauthor   = {Bicheng Wang, Xinyi Zhang},
  pdfsubject  = {Quantitative finance; machine learning},
  pdfkeywords = {deep reinforcement learning, algorithmic trading, actor-critic methods, portfolio management, backtesting}
}

\setlist{itemsep=2pt,topsep=4pt}
\newcommand{\ind}[1]{\texttt{\detokenize{#1}}}   
\newcommand{\bestcell}[1]{\textbf{#1}}

\title{\vspace{-0.6in}\textbf{Deep Reinforcement Learning for Equity Trading:\\
Benchmarking Actor-Critic Methods with Forward Retraining}}

\author{
  \begin{tabular}{c@{\hspace{3em}}c}
    \textbf{Bicheng Wang} & \textbf{Xinyi Zhang} \\
    Stanford University   & Stanford University  \\
    \texttt{bichengw@stanford.edu} & \texttt{xyzh@stanford.edu}
  \end{tabular}
}

\date{June 8, 2021}

\begin{document}
\maketitle

\begin{abstract}
Consistently profitable trading is difficult because equity markets are noisy, non-stationary, and only partially predictable from historical data. We benchmark five deep reinforcement learning (DRL) actor-critic methods---A2C, PPO, DDPG, TD3, and SAC---that learn trading actions end to end from market states, and compare them with a supervised price-forecasting baseline. Using daily data for 20 large-capitalization S\&P~500 stocks from 2000 to 2020, enriched with trend-following technical indicators and log min--max scaling, we train on 2000--2018 and backtest on 2019--2020. Each agent is evaluated both when trained once and under forward retraining, in which it is retrained on all data available before each successive test window. DDPG achieves the highest annual return (55.5\%), Sharpe ratio (1.38), and alpha (0.22), but also the highest market beta (1.24). TD3 and SAC offer a better risk--return balance, with Sharpe ratios of 1.37 and 1.33 and maximum drawdowns of about 25\%. Forward retraining improves A2C, PPO, and SAC, leaves TD3 essentially unchanged, and reduces DDPG's annual return from 55.5\% to 29.8\%, consistent with TD3's greater robustness to hyperparameters. The forecasting baseline has the smallest maximum drawdown (9.6\%) and the lowest beta (0.31), underscoring a trade-off between the higher returns of end-to-end DRL and the lower risk of forecast-driven strategies.
\end{abstract}


\section{Introduction}
\label{sec:intro}

Profitable trading is central to investment management. Because the stock market is dynamic and complex, it is difficult to profit from trading consistently. Machine learning offers clear advantages in data mining, forecasting, and automated trading, and in this work we explore how deep learning can be used to construct a profitable stock portfolio.

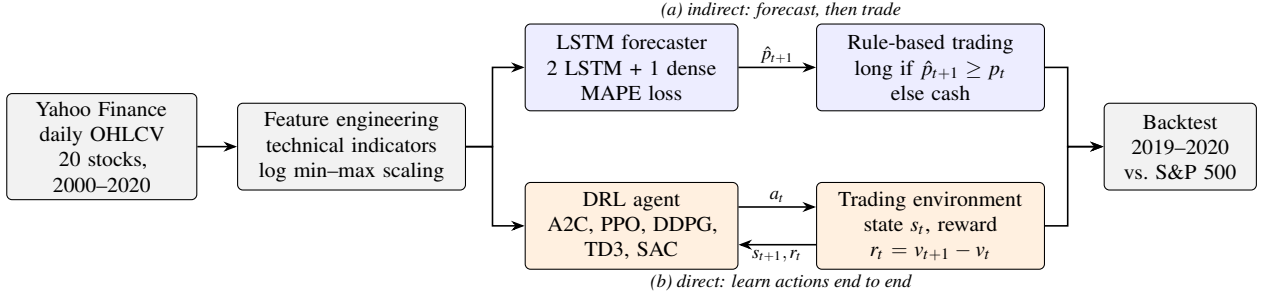
\begin{figure}[t]
\centering
\begin{tikzpicture}[
  font=\footnotesize,
  box/.style={draw, rounded corners=2pt, align=center, minimum height=1.15cm, inner sep=3pt},
  data/.style={box, fill=gray!10},
  lstm/.style={box, fill=blue!7},
  rl/.style={box, fill=orange!12},
  arr/.style={-{Stealth[length=1.8mm]}, semithick},
  lab/.style={font=\scriptsize, inner sep=1.5pt}
]
  \node[data, text width=2.3cm] (raw)   at (1.25, 0)    {Yahoo Finance\\daily OHLCV\\20 stocks, 2000--2020};
  \node[data, text width=2.8cm] (feat)  at (4.55, 0)    {Feature engineering\\technical indicators\\log min--max scaling};
  \node[lstm, text width=2.6cm] (lstm)  at (8.25, 1.05) {LSTM forecaster\\2 LSTM + 1 dense\\MAPE loss};
  \node[lstm, text width=2.8cm] (rule)  at (12.2, 1.05) {Rule-based trading\\long if $\hat{p}_{t+1} \ge p_t$\\else cash};
  \node[rl,   text width=2.6cm] (agent) at (8.25,-1.05) {DRL agent\\A2C, PPO, DDPG,\\TD3, SAC};
  \node[rl,   text width=2.8cm] (env)   at (12.2,-1.05) {Trading environment\\state $s_t$, reward\\$r_t = v_{t+1} - v_t$};
  \node[data, text width=1.8cm] (bt)    at (15.5, 0)   {Backtest\\2019--2020\\vs.\ S\&P~500};

  \draw[arr] (raw) -- (feat);
  \draw[arr] (feat.east) -- ++(0.35,0) |- (lstm.west);
  \draw[arr] (feat.east) -- ++(0.35,0) |- (agent.west);
  \draw[arr] (lstm) -- node[lab, above] {$\hat{p}_{t+1}$} (rule);
  \draw[arr] ([yshift=2.5mm]agent.east) -- node[lab, above] {$a_t$} ([yshift=2.5mm]env.west);
  \draw[arr] ([yshift=-2.5mm]env.west) -- node[lab, below] {$s_{t+1}, r_t$} ([yshift=-2.5mm]agent.east);
  \draw[arr] (rule.east) -- ++(0.3,0) |- (bt.west);
  \draw[arr] (env.east)  -- ++(0.3,0) |- (bt.west);

  \node[lab, font=\scriptsize\itshape, anchor=south] at ($(lstm.north)!0.5!(rule.north)$) {(a) indirect: forecast, then trade};
  \node[lab, font=\scriptsize\itshape, anchor=north] at ($(agent.south)!0.5!(env.south)$) {(b) direct: learn actions end to end};
\end{tikzpicture}
\caption{Overview of the two approaches. (a) An LSTM forecasts prices and a rule-based strategy turns the forecasts into positions. (b) A DRL agent interacts with a trading environment and learns trading actions end to end. Both are trained on 2000--2018 and backtested on 2019--2020.}
\label{fig:overview}
\end{figure}

We study two complementary approaches (\cref{fig:overview}). The \emph{indirect} approach uses time-series forecasting: an LSTM network predicts the market price, and a basic trading strategy acts on the prediction. The \emph{direct} approach uses deep reinforcement learning (DRL): an agent observes the market and outputs trading actions end to end to build the portfolio. Our contributions are as follows:
\begin{itemize}
  \item We construct a dataset of 20 large-capitalization S\&P~500 stocks over 2000--2020, enriched with trend-following technical indicators and log min--max scaling (\cref{sec:data}).
  \item We study loss functions, architectures, and training-data regimes for LSTM price forecasting, and find that a MAPE loss gives the lowest validation error among the losses we compare (\cref{sec:lstm}).
  \item We benchmark five DRL algorithms (A2C, PPO, DDPG, TD3, and SAC), each trained once and with forward retraining, against the LSTM strategy on a 2019--2020 out-of-sample backtest (\cref{sec:drl,sec:results}).
\end{itemize}

\section{Related Work}
\label{sec:related}

Our use of LSTM networks~\citep{hochreiter1997lstm} to predict stock prices was inspired by~\citep{sayah2020kaggle}. \citet{moghar2020stock} summarize design experience with LSTM recurrent networks for stock market prediction, and the TensorFlow time-series tutorial~\citep{tensorflow2020timeseries} provides an architectural reference for forecasting models.

Reinforcement learning for equity trading has attracted growing attention in recent years~\citep{yang2020ensemble,fischer2018survey,wang2019alphastock,dai2019fx,xiong2018practical,theate2021application}, including ensemble DRL strategies for automated stock trading~\citep{yang2020ensemble}, surveys of RL in financial markets~\citep{fischer2018survey}, and interpretable attention-based portfolio agents~\citep{wang2019alphastock}. Several challenges remain. (i)~Real-world trading data are limited. (ii)~The reward of a trading strategy can be defined in many ways, requiring trade-offs among a robust learning rule, the final optimization target, and the limitations of the dataset. (iii)~Data sparsity~\citep{nasiri2017sparsity}: because training relies solely on historical data, patterns that occurred rarely in the past may not be captured.

\section{Dataset and Features}
\label{sec:data}

We select the 20 stocks with the largest market capitalization among S\&P~500 constituents and collect their daily data from 2000 to 2020. We restrict the universe to these stocks because long, complete historical records are limited and these companies account for a large share of the index's total market capitalization. Companies that went public after 2000 are excluded so that every stock has complete data over the entire period. Raw data are obtained from the Yahoo Finance API. Each record contains the date, open, high, low, and close prices, trading volume, ticker symbol, and day of the week. The dataset contains 105{,}680 rows in total (20 stocks $\times$ 5{,}284 trading days). We split the data chronologically into a training set (2000--2018) and a test set (2019--2020), approximately a 90/10 split. \Cref{tab:raw} shows sample records.

\begin{table}[ht]
\centering
\small
\caption{Sample of the raw data. \emph{Close} is the split- and dividend-adjusted closing price, which is why it can lie outside the unadjusted low--high range.}
\label{tab:raw}
\begin{tabular}{lrrrrrl}
\toprule
Date & Open & High & Low & Close & Volume & Ticker \\
\midrule
2000-01-03 &  0.936384 &  1.004464 &  0.907924 &  0.859423 & 535{,}796{,}800 & AAPL  \\
2000-01-03 & 16.812500 & 16.875000 & 16.062500 & 16.274673 &   7{,}384{,}400 & ADBE  \\
2000-01-03 & 81.500000 & 89.562500 & 79.046875 & 89.375000 &  16{,}117{,}600 & AMZN  \\
2000-01-03 & 25.125000 & 25.125000 & 24.000000 & 13.952057 &  13{,}705{,}800 & BAC   \\
2000-01-03 & 36.500000 & 36.580002 & 34.820000 & 35.299999 &     875{,}000   & BRK-B \\
\bottomrule
\end{tabular}
\end{table}

Feature engineering is a crucial step in training a high-quality model: it converts the raw data into a model-ready state. We first check for missing data and then construct the features described below.

\subsection{Technical Indicators}
\label{sec:indicators}

In practice, a trader takes many sources of information into account, such as historical prices, current holdings, and technical indicators. We add seven trend-following technical indicators: moving average convergence divergence (MACD), relative strength index (RSI), Bollinger bands (BOLL), commodity channel index (CCI), directional movement index (DX), simple moving average (SMA), and exponential moving average (EMA), several of them computed over multiple look-back windows (see \cref{sec:mdp} for the full list). \Cref{tab:features} shows a sample of the processed dataset.

\begin{table}[ht]
\centering
\small
\caption{Sample of the processed dataset (first trading day; intermediate columns omitted).}
\label{tab:features}
\begin{tabular}{lcccccccc}
\toprule
Ticker & Day & \ind{macd} & \ind{boll_ub} & $\cdots$ & \ind{cci_10} & \ind{dx_30} & \ind{close_120_sma} & \ind{close_120_ema} \\
\midrule
AAPL  & 0 & 0 & 0.9256 & $\cdots$ & $-66.67$ & 100 &  0.859 &  0.859 \\
ADBE  & 0 & 0 & 0.9256 & $\cdots$ & $-66.67$ & 100 & 16.274 & 16.274 \\
AMZN  & 0 & 0 & 0.9256 & $\cdots$ & $-66.67$ & 100 & 89.375 & 89.375 \\
BAC   & 0 & 0 & 0.9256 & $\cdots$ & $-66.67$ & 100 & 13.952 & 13.952 \\
BRK-B & 0 & 0 & 0.9256 & $\cdots$ & $-66.67$ & 100 & 35.299 & 35.299 \\
\bottomrule
\end{tabular}
\end{table}

\subsection{Logarithmic Scaling}
\label{sec:logscale}

Over the long run, the U.S.\ equity market has grown at a roughly exponential rate despite intermittent recessions, as the century-long history of the S\&P~500 illustrates. Exponential growth introduces nonlinearity into price-level features, and features such as the price itself, SMAs, and EMAs are better modeled on a logarithmic scale. We therefore apply log min--max scaling to these features,
\begin{equation}
  \tilde{x} = \frac{\log x - \min(\log x)}{\max(\log x) - \min(\log x)},
  \label{eq:logscale}
\end{equation}
which makes their patterns easier for a relatively shallow network to capture.

\section{Methods}
\label{sec:methods}

We investigate two approaches to optimizing stock trading strategies. We first combine LSTM time-series forecasting with a simple single-stock trading strategy, and then use DRL models to optimize trading directly.

\subsection{LSTM Time-Series Forecasting}
\label{sec:lstm}

The LSTM approach first forecasts the market and then applies a simple strategy to build a portfolio. We designed this forecasting-and-trading pipeline end to end.

\subsubsection{Loss Function}
\label{sec:loss}

The choice of loss function and evaluation metric strongly influences what a model ultimately optimizes, and for stock prediction the appropriate loss differs from that of a generic regression problem. We want predictions to be close to the true values, but ``close'' must be measured appropriately. Consider AAPL in 2000, when a share was worth about \$1: a \$0.10 increase is a 10\% return. In 2021, the equivalent event is a \$200 share rising to \$220. Mean squared error (MSE) on price levels weights the second error $(20/0.1)^2 = 40{,}000$ times more heavily than the first, even though both represent the same 10\% move. MSE therefore underweights low-price periods and produces an unbalanced fit. We compared MSE, mean absolute percentage error (MAPE), mean squared logarithmic error (MSLE), Huber loss, and log-cosh loss.

MAPE better captures the relationship between predicted and actual prices over long periods in which price levels change substantially; in our experiments, models trained with MAPE showed less prediction skew as prices rose. With actual prices $y_i$ and predictions $\hat{y}_i$,
\begin{equation}
  \mathcal{L}_{\text{MAPE}} = \frac{100}{n} \sum_{i=1}^{n} \left| \frac{\hat{y}_i - y_i}{y_i} \right|,
  \label{eq:mape}
\end{equation}
whose derivative with respect to each prediction is
\begin{equation}
  \frac{\partial \mathcal{L}_{\text{MAPE}}}{\partial \hat{y}_i} =
  \begin{cases}
    -\dfrac{100}{n\,|y_i|} & \text{if } \hat{y}_i < y_i, \\[6pt]
    \text{undefined}        & \text{if } \hat{y}_i = y_i, \\[2pt]
    \phantom{-}\dfrac{100}{n\,|y_i|} & \text{if } \hat{y}_i > y_i.
  \end{cases}
  \label{eq:mape-grad}
\end{equation}
Each sample therefore contributes a gradient of constant magnitude that is inversely proportional to its price level, so low-price periods are not underweighted. A squared percentage error, whose gradient scales with the size of the error, is a natural variant when smoother gradients near the optimum are desired.

\subsubsection{Architecture and Hyperparameters}
\label{sec:arch}

\paragraph{Architecture.} We explored bidirectional LSTMs, stacked LSTMs with additional layers, and LSTMs followed by multiple dense layers of various sizes. A two-layer LSTM followed by a single dense layer yielded the most stable predictive model.

\paragraph{Units and dropout.} At our dataset scale, too many parameters lead to overfitting; mitigating this with dropout or other regularization increased computational cost without improving the metrics (\cref{tab:lstm}).

\paragraph{Training data.} With 20 stocks, several training regimes are possible: (i)~train and predict on a single stock; (ii)~train on all stocks and predict a single stock; and (iii)~train on all stocks, fine-tune on the target stock, and then predict. Training on all stocks already yields good predictions (compare \emph{Baseline} with \emph{Single ticker} in \cref{tab:lstm}), although the model can optionally be fine-tuned on a specific stock before prediction.

\Cref{tab:lstm} compares the models we explored.

\begin{table}[t]
\centering
\small
\caption{Training and validation metrics of the LSTM variants. \emph{Loss} is each model's training objective, which is MAPE except for the MSE, Huber, and log-cosh models. Metrics are computed on scaled prices; the subscript \emph{val} denotes validation. Best value in each metric column in bold.}
\label{tab:lstm}
\begin{tabular}{lrrrrrrr}
\toprule
Model & Loss & MAE & MAPE & MSE ($\times10^{-6}$) & MAE$_{\text{val}}$ & MAPE$_{\text{val}}$ & MSE$_{\text{val}}$ ($\times10^{-6}$) \\
\midrule
Baseline      & 1.052367 & 0.004578 & 1.052367 & 38 & 0.004099 & 0.972073 & 32 \\
Dropout       & 2.371855 & 0.011330 & 2.371855 & 271 & 0.007442 & 1.535101 & 95 \\
MSE           & 0.000336 & 0.006153 & 1.412530 & 336 & 0.004355 & 1.036787 & 38 \\
Huber         & 0.000098 & 0.005407 & 1.253396 & 196 & 0.004077 & 0.974119 & 35 \\
Log-cosh      & 0.000066 & 0.005286 & 1.226172 & 133 & 0.003798 & 0.898953 & 29 \\
MAPE          & 0.935923 & \bestcell{0.004015} & \bestcell{0.935923} & \bestcell{31} & 0.003789 & 0.888290 & 29 \\
Single ticker & 2.464035 & 0.010135 & 2.464035 & 172 & 0.010011 & 2.582896 & 145 \\
Fewer units   & 0.995828 & 0.004298 & 0.995828 & 35 & \bestcell{0.003565} & \bestcell{0.856291} & \bestcell{27} \\
More units    & 1.074033 & 0.004663 & 1.074033 & 39 & 0.005058 & 1.143134 & 44 \\
\bottomrule
\end{tabular}
\end{table}

\subsubsection{Trading Strategy}
\label{sec:strategy}

Given the LSTM forecasts, we apply a simple modified buy-and-hold strategy. If the predicted price is lower than the current price, the strategy holds cash; if it is higher than or equal to the current price, it buys shares and holds them. Formally, the position on day $t$ is
\begin{equation}
  w_t = \lambda \cdot \mathbb{1}\!\left[\hat{p}_{t+1} \ge p_t\right],
  \label{eq:strategy}
\end{equation}
where $p_t$ is the current price, $\hat{p}_{t+1}$ is the predicted price, and $\lambda = 1$ for the unleveraged strategy and $\lambda > 1$ for the leveraged one. We evaluate both variants (\cref{tab:results}).

\subsection{Deep Reinforcement Learning}
\label{sec:drl}

The DRL approach aims to generate portfolio trading actions directly, end to end, from the market environment. We formulate trading as a Markov decision process.

\subsubsection{Model Definition}
\label{sec:mdp}

\paragraph{Action.} The action space describes how the agent interacts with the environment. For a single stock, a basic action takes one of three values, $a \in \{-1, 0, 1\}$, representing selling, holding, and buying one share. More generally, an action may involve multiple shares, so for each stock we use the action space $a \in \{-k, \dots, -1, 0, 1, \dots, k\}$, where $k$ is the maximum number of shares per trade. For example, ``buy 10 shares of AAPL'' and ``sell 10 shares of AAPL'' correspond to $a = 10$ and $a = -10$, respectively.

\paragraph{Reward.} The reward function $r(s, a, s')$ is the incentive for the agent to learn better actions. It is the change in portfolio value when action $a$ is taken in state $s$, leading to the new state $s'$:
\begin{equation}
  r(s, a, s') = v' - v,
  \label{eq:reward}
\end{equation}
where $v$ and $v'$ are the portfolio values at states $s$ and $s'$, respectively.

\paragraph{State.} The state space describes the observations the agent receives from the environment. Just as a human trader analyzes many sources of information before executing a trade, our agent observes a range of features in order to learn in the interactive environment. The state is $s = [\,b,\ p,\ h,\ \mathbf{x}\,]$, where $b$ is the available cash balance, $p$ the closing prices, $h$ the number of shares held of each stock, and $\mathbf{x}$ the technical indicators of \cref{sec:indicators}: \ind{macd}, \ind{boll_ub}, \ind{boll_lb}, \ind{rsi_10}, \ind{rsi_20}, \ind{cci_10}, \ind{cci_20}, \ind{dx_30}, \ind{close_20_sma}, \ind{close_60_sma}, \ind{close_120_sma}, \ind{close_20_ema}, \ind{close_60_ema}, and \ind{close_120_ema}. The numeric suffix of each indicator denotes its look-back window in trading days.

\subsubsection{Learning Algorithms}
\label{sec:algos}

We evaluate five DRL algorithms.

\paragraph{Proximal Policy Optimization (PPO).} PPO~\citep{schulman2017ppo} is a policy-gradient method that limits the size of each policy update. It simplifies trust region policy optimization (TRPO)~\citep{schulman2015trpo}, which enforces a hard constraint on the KL divergence between successive policies, by instead optimizing a clipped surrogate objective (or, in a variant, adding a KL-divergence penalty). This keeps policy learning stable using only first-order optimization. Because each batch of collected experience is reused for several epochs of minibatch updates, PPO is relatively sample efficient, which is valuable when historical trading data are scarce.

\paragraph{Advantage Actor-Critic (A2C).} Whereas PPO optimizes the policy directly, value-based methods, which estimate the expected return, can also improve learning. Actor-critic methods~\citep{konda2000actor,pfau2016connecting} combine the two: an actor updates the policy while a critic estimates the value function. We use A2C, the synchronous variant of A3C~\citep{mnih2016a3c}, whose advantage estimate $A(s,a) = r + \gamma V(s') - V(s)$ reduces the variance of policy-gradient updates.

\paragraph{Deep Deterministic Policy Gradient (DDPG).} DDPG~\citep{lillicrap2016ddpg} is an off-policy actor-critic algorithm designed for continuous action spaces, which suits the large, ordered share-count action space. It concurrently learns a Q-function and a deterministic policy: the Q-function is learned from off-policy data via the Bellman equation, and the policy is learned by maximizing the Q-function. Our DDPG implementation builds on Stable-Baselines3~\citep{raffin2019sb3}.

\paragraph{Twin Delayed DDPG (TD3).} In our experiments, DDPG sometimes performed very well but was frequently brittle with respect to hyperparameters and other tuning choices, which caused instability. TD3~\citep{fujimoto2018td3} addresses these issues with three modifications: it learns two Q-functions and uses the smaller of the two Q-values to form the targets in the Bellman error loss (clipped double-Q learning); it updates the policy less frequently than the Q-functions (delayed policy updates); and it adds noise to the target action to smooth the Q-function (target policy smoothing).

\paragraph{Soft Actor-Critic (SAC).} SAC~\citep{haarnoja2018sac} optimizes a stochastic policy in an off-policy manner, bridging stochastic policy optimization and DDPG-style approaches. Its central feature is entropy regularization: the policy is trained to maximize a trade-off between expected return and entropy, which encourages exploration and helps prevent premature convergence to a poor local optimum.

\subsubsection{Forward Retraining}
\label{sec:rolling}

We compare agents trained once on the training period with agents under \emph{forward retraining}. In forward retraining, the test period is divided into consecutive time slices; before each slice, the agent is retrained on all historical data available up to that point and then trades during that slice. This is a walk-forward scheme with an expanding training window: the model never sees data from the slice it trades.

\section{Results}
\label{sec:results}

\subsection{Evaluation Metrics}

We evaluate each strategy on the 2019--2020 test period using annual return, cumulative return~\citep{cumret_investopedia}, Sharpe ratio~\citep{sharpe_investopedia}, maximum drawdown~\citep{mdd_investopedia}, alpha, and beta, with the S\&P~500 index as the benchmark. The Sharpe ratio measures return per unit of volatility; maximum drawdown is the largest peak-to-trough decline in portfolio value; and alpha and beta are the intercept and slope of a regression of strategy returns on benchmark returns, measuring excess return and market exposure, respectively. \Cref{tab:results} reports the results.

\begin{table}[t]
\centering
\small
\caption{Backtest performance on the 2019--2020 test set; the benchmark is the S\&P~500. \emph{Retrained} denotes forward retraining (\cref{sec:rolling}); other DRL rows are trained once. Best value in each column in bold (for maximum drawdown, the smallest decline); beta is a measure of market exposure and is not ranked.}
\label{tab:results}
\begin{tabular}{lrrrrrr}
\toprule
Model & Annual return & Cumulative return & Sharpe ratio & Max drawdown & Alpha & Beta \\
\midrule
LSTM               & 16.99\% &  36.85\% & 1.19 & \bestcell{$-9.57\%$} & 0.09 & 0.31 \\
LSTM (leveraged)   & 34.25\% &  80.22\% & 1.19 & $-19.14\%$ & 0.20 & 0.61 \\
\midrule
A2C                & 20.47\% &  45.23\% & 0.82 & $-31.11\%$ & $-0.03$ & 1.02 \\
A2C (retrained) & 29.83\% &  68.73\% & 0.99 & $-33.52\%$ & 0.03 & 1.14 \\
\midrule
PPO                & 27.89\% &  63.71\% & 1.09 & $-25.33\%$ & 0.05 & 0.91 \\
PPO (retrained) & 32.16\% &  74.84\% & 1.30 & $-22.84\%$ & 0.10 & 0.84 \\
\midrule
DDPG               & \bestcell{55.51\%} & \bestcell{142.26\%} & \bestcell{1.38} & $-33.52\%$ & \bestcell{0.22} & 1.24 \\
DDPG (retrained) & 29.82\% &  68.72\% & 1.09 & $-33.52\%$ & 0.04 & 1.02 \\
\midrule
TD3                & 42.00\% & 101.93\% & 1.37 & $-25.24\%$ & 0.18 & 0.89 \\
TD3 (retrained) & 41.89\% & 101.61\% & 1.27 & $-28.68\%$ & 0.13 & 1.11 \\
\midrule
SAC                & 38.95\% &  93.32\% & 1.33 & $-24.19\%$ & 0.15 & 0.90 \\
SAC (retrained) & 40.44\% &  97.51\% & 1.28 & $-23.20\%$ & 0.19 & 0.83 \\
\bottomrule
\end{tabular}
\end{table}

\subsection{Discussion}

\paragraph{LSTM.} The unleveraged LSTM strategy has the lowest annual return (16.99\%), but its Sharpe ratio of 1.19 lies in the middle of the range, and it has the smallest maximum drawdown ($-9.57\%$) and the lowest beta (0.31) of all strategies. Leverage roughly doubles its annual return, maximum drawdown, alpha, and beta while leaving the Sharpe ratio unchanged. The resulting alpha of 0.20 is second only to DDPG (0.22), and the beta of 0.61 remains well below that of every DRL agent (0.83--1.24). Because the LSTM's price forecasts are easy for humans to interpret, this approach also has strong potential to be combined with other trading strategies.

\paragraph{DRL agents.} DDPG attains the highest annual return (55.51\%), cumulative return (142.26\%), Sharpe ratio (1.38), and alpha (0.22), but also the highest beta (1.24) and one of the deepest drawdowns ($-33.52\%$). TD3 and SAC offer a better balance between return and risk, with Sharpe ratios of 1.37 and 1.33 and maximum drawdowns of $-25.24\%$ and $-24.19\%$, respectively.

\paragraph{Forward retraining.} Forward retraining improves the annual return of A2C (20.47\% $\to$ 29.83\%), PPO (27.89\% $\to$ 32.16\%), and SAC (38.95\% $\to$ 40.44\%), and leaves TD3 essentially unchanged (42.00\% vs.\ 41.89\%). DDPG is the exception: its annual return falls from 55.51\% to 29.82\%. In our experience, DDPG is difficult to tune to a suitable configuration for every retraining window, which may explain the gap between the individually tuned, once-trained DDPG and its retrained counterpart.

\paragraph{DDPG vs.\ TD3.} Consistent with this observation, TD3 substantially stabilized training in our experiments, making results far less dependent on hyperparameters and tuning, which is a significant advantage in a market that is hard to predict. Its performance is nearly identical with and without forward retraining.

\section{Conclusion}
\label{sec:conclusion}

We applied an LSTM time-series forecasting model and five deep reinforcement learning algorithms to learn profitable trading strategies for large-capitalization U.S.\ equities. For the LSTM, we explored loss functions, architectures, and training regimes to select the best model, and found MAPE to be the most effective loss among those compared. For DRL, we evaluated A2C, PPO, DDPG, TD3, and SAC under two training strategies. On the 2019--2020 test period, the DRL agents achieved the highest returns, led by DDPG with a 55.51\% annual return; the LSTM strategy offered the smallest drawdown and market exposure; and TD3 combined strong risk-adjusted performance with stability across training strategies.

\paragraph{Limitations.} Our stock universe consists of large companies with complete 2000--2020 histories, which introduces survivorship bias. The test period covers only two years (2019--2020), which include both a strong bull market and the COVID-19 crash of early 2020, so the results may not generalize to other market regimes. Finally, DRL performance is known to be sensitive to random seeds and hyperparameters, and we do not report variability across runs. Future work could address these issues with a point-in-time stock universe, longer walk-forward evaluation, and multiple random seeds.

\section*{Author Contributions}

Bicheng Wang developed the LSTM forecasting model and half of the reinforcement learning models. Xinyi Zhang developed the other half of the reinforcement learning models.

\section*{Acknowledgments}

This work was carried out under the guidance of Ayush Kanodia, who advised on comparing the LSTM and reinforcement learning approaches and on the forward-retraining scheme. We thank Huizi Mao for discussions that inspired the original trading-model design. We also thank Andrew Ng and Kian Katanforoosh, the instructors of Stanford CS230, whose course informed our choices of LSTM hyperparameters, loss functions and their derivatives, evaluation metrics, and reinforcement learning methods.

\bibliographystyle{unsrtnat}
{\small
\setlength{\bibsep}{3pt plus 1pt}
\bibliography{references}
}

\end{document}